\documentclass[twoside]{article}

\usepackage{aistats2020}
\usepackage[utf8]{inputenc} 
\usepackage[T1]{fontenc}    
\usepackage{hyperref}       
\usepackage{url}            
\usepackage{booktabs}       
\usepackage{nicefrac}       
\usepackage{microtype}   
\usepackage{bm}
\usepackage{amsfonts,amsmath,amssymb,amsthm,bm, mathrsfs}
\usepackage{subcaption} 
\usepackage{graphicx} 

\newtheorem{theorem}{Theorem}[section]

\newtheorem{proposition}[theorem]{Proposition}
\newtheorem{definition}[theorem]{Definition}
\newtheorem{example}[theorem]{Example}

\newcommand{\T}{\mathbb{T}}
\newcommand{\hilb}{\mathcal{H}}
\newcommand{\X}{\mathcal{X}}
\newcommand{\I}{\mathcal{I}}
\newcommand{\B}{\mathcal{B}}

\begin{document}
%

%

\twocolumn[

\aistatstitle{Metric on Random Dynamical Systems with Vector-valued Reproducing Kernel Hilbert Spaces}


\aistatsauthor{  Isao Ishikawa \\ isao.ishikawa@riken.jp 
\And Akinori Tanaka \\ akinori.tanaka@riken.jp 
\And Masahiro Ikeda \\masahiro.ikeda@riken.jp
\And Yoshinobu Kawahara \\kawahara@imi.kyushu-u.ac.jp}

\aistatsaddress{ RIKEN  / Keio University
\And RIKEN  / Keio University
\And RIKEN  / Keio University
\And Kyushu University} 
]

\begin{abstract}
Development of metrics for structural data-generating mechanisms is fundamental in machine learning and the related fields. 
In this paper, we give a general framework to construct metrics on {\em random} nonlinear dynamical systems, defined with the Perron-Frobenius operators in vector-valued reproducing kernel Hilbert spaces (vvRKHSs). 
We employ vvRKHSs to design mathematically manageable metrics and also to introduce 
operator-valued kernels, which enables us to handle randomness in systems. 
Our metric provides an extension of the existing metrics for {\em deterministic} systems, and gives a specification of the kernel maximal mean discrepancy of random processes.
Moreover, by considering the time-wise independence of random processes, we clarify a connection between our metric and the independence criteria with kernels such as Hilbert-Schmidt independence criteria. 
We empirically illustrate our metric with synthetic data, and evaluate it in the context of the independence test for random processes. We also evaluate the performance with real time seris datas via clusering tasks.
\end{abstract}

\section{Introduction}
%
%
Development of a metric for data-generating mechanisms is fundamental in machine learning and the related fields. This is because the development of an algorithm for respective learning problems according to the type of data structures in many cases is basically reduced to the design of an appropriate metric or kernel. As for the context of dynamical system, the majority of the existing metrics for dynamical systems have been developed with principal angles between some appropriate subspaces such as column subspaces of observability matrices \cite{Martin00,DeCock-DeMoor02,VSV07}. Recently, several metrics on dynamical systems are developed with transfer operators such as Koopman operators and Perron-Frobenius operators \cite{KSM19}. Mezic~et~al.~\cite{MB04,Mezic16} propose metrics of dynamical systems in the context of the ergodic theory via Koopman operators on $L^2$-spaces. Fujii~et~al.~\cite{Fujii17} developed metrics with Koopman operators as a generalization of the ones with Binet-Cauchy kernel proposed by Vishwanathan~et~al.~\cite{VSV07}. And, Ishikawa~et~al.~\cite{IFI+18} gave metrics on nonlinear dynamical systems with Perron-Frobenius operators in RKHSs, which generalize the classical ones with principal angles mentioned above.

However, the above existing metrics are basically defined for {\em deterministic} dynamical systems. And, to the best of our knowledge, few existing literature has addressed the design of metrics for {\em random} dynamical systems or stochastic processes despite its importance for data analysis. Vishwanathan~et~al.~\cite{VSV07} mentioned their metrics for cases where systems include random noises by taking expectations over the randomness. And, Chwialkowski and Gretton ~\cite{CG14} developed non-parametric test statistics for random processes by extending the Hilbert Schmidt independence criteria.




In this paper, we give a framework to construct metrics on {\em random} nonlinear dynamical systems, which are defined with the Perron-Frobenius operators in vector-valued reproducing kernel Hilbert spaces (vvRKHSs). Here, we employ vvRKHSs to design mathematically manageable metrics and also to introduce 
operator-valued kernels, which enable us to handle the randomness in systems. We first define the Perron-Frobenius operators in vvRKHSs and construct a dynamical system in a canonical way. And based on these, we define a metric on random dynamical systems as a positive definite operator-valued kernel. Our metric provides an extension of the existing metrics for deterministic systems to random ones, and gives a specification of the kernel maximal mean discrepancy of random processes. Moreover, by considering the time-wise independence of random processes, we clarify a connection between our metric and the independence criteria with kernels such as Hilbert-Schmidt independence criteria. We empirically illustrate our metric using synthetic data, and evaluate it in the context of the independence test for random processes.

The remainder of this paper is orgnized as follows. In Section~\ref{sec:background}, we first briefly review the notions necessary in this paper such as the Perron-Frobenius operators in RKHSs, and a positive definite kernel on random processes by means of the kernel mean embeddings. In Section~\ref{sec:r2d}, we construct a vvRKHS to treat random dynamical system, and define the Perron-Frobenius operators on the vvRKHSs. Then, in Section~\ref{sec:metric}, we give the definition of our metric for random dynamical systems. In Section~\ref{sec:connection}, we describe the connection of our metric to the Hilbert Schmidt independence criteria. Finally, we investigate empirically our metric using both synthetic and real data in Section~\ref{sec:experiment}, and then conclude the paper in Section~\ref{sec:concl}. All proofs are given in Appendix~A.
\section{Background}
\label{sec:background}

In this section, we briefly review the Perron-Frobenius operators in RKHSs in Subsection~\ref{ssec:pf}, and we introduce the notion of random dynamical systems in Subsection~\ref{ssec:rds}.  In the end, we describe a  way of defining a metric for comparing two random processes with the kernel mean embeddings in Subsection~\ref{ssec:comp_rp}.

\subsection{Perron-Frobenius Opertors in RKHSs}
\label{ssec:pf}

Let $\mathcal{X}$ be a state space and $k$ be a positive definite kernel on $\mathcal{X}$.  For any $a\in\mathcal{X}$, we denote by $k_a$ the function on $\mathcal{X}$ defined by $k_a(x)=k(a,x)$.  By Moore-Aronszajn' theorem, there exists a unique Hilbert space $\hilb_k$ composed of functions of $\mathcal{X}$ such that for any $a\in \mathcal{X}$, the function $k_a$ is contained in $\hilb_k$ and the reproducing property holds, namely, for any $f\in\hilb_k$, $\langle f, k_a\rangle_{\hilb_k}=f(a)$. The Gaussian kernel $k(x,y)=e^{-|x-y|^2}$ for $x,y\in\mathbb{R}^d$ is a typical example of the positive definite kernel.

Let $\T:=\mathbb{Z}$, $\mathbb{Z}_{\ge0}$, $\mathbb{R}$, or $\mathbb{R}_{\ge0}$.  
We call a map $\varphi:\T \times \mathcal{X} \rightarrow \mathcal{X}$ a {\em dynamical system} if $\varphi(0,x)=x$ and for any $s,t\in\T$ and $x\in\mathcal{X}$ , $\varphi(s+t, x)=\varphi(s,\varphi(t, x))$.   
For $t\in\T$, we define the {\em Perron-Frobenius operator} $K_\varphi^t:\hilb_k\rightarrow\hilb_k$ by a linear operator with a dense domain, ${\rm span}\{k_x~|~x\in\mathcal{X}\}$, by
\begin{equation*}
K_\varphi^tk_x:=k_{\varphi(t,x)}.
\end{equation*}
As in the same manner as Proposition~2.1 in \cite{IFI+18}, $K_\varphi^t$ is the adjoint operator of the Koopman operator in $\hilb_k$, which is a linear operator allocating $g\in\hilb_k$ to $g(\varphi(t,\cdot))$. Note that, although the contents in \cite{IFI+18} are considered only for the discrete time case, i.e., $\T=\mathbb{Z}$ or $\mathbb{Z}_{\ge0}$, we can consider the general case $\T$. Ishikawa~et~al.~\cite{IFI+18} define a positive definite kernel by using the Perron-Frobenius operators $K_\varphi^t$ for comparing {\em deterministic} nonlinear dynamical systems, which generalizes many of the existing metrics for dynamical systems \cite{Martin00,DeCock-DeMoor02,VSV07}.


\subsection{Random Dynamical Systems (RDS)}
\label{ssec:rds}
Let $(\Omega, P)$ be a probability space, where $\Omega$ is a measurable space, and $P$ is a probability measure. Let 
$\T:=\mathbb{Z}$, $\mathbb{Z}_{\ge0}$, $\mathbb{R}$, or $\mathbb{R}_{\ge0}$, 
and $\mathcal{X}$ be a state space. 
We fix a semi-group of measure preserving measurable maps $\Theta:=\{\theta_t\}_{t\in \T}$ on $\Omega$, namely, $\theta_t: \Omega\rightarrow\Omega$ such that $(\theta_t)_*P=P$, $\theta_0={\rm id}_{\Omega}$, and $\theta_s\circ\theta_t=\theta_{s+t}$ for all $s,t\in\T$. Here, $(\theta_t)_*P$ is the push-forward measure of $P$. 
\begin{definition}[Random dynamical system]

Let $\mathcal{M}\subset\mathcal{X}$ be an open subset. A {\em random dynamical system} on $\Omega$ with respect to $\Theta
$ is a measurable map
\[\Phi: \T\times\Omega\times\mathcal{M}\rightarrow\mathcal{M}\]
such that $\Phi(0, \omega, x)=x$ and $\Phi(t+s, \omega, x)=\Phi\big(t,\theta_s(\omega),\Phi(s,\omega, x)\big)$ for any $x\in\mathcal{M}$. 
\end{definition}
Random dynamical systems include many kinds of stochastic processes, and typically appear as solutions of stochastic differential equations. In the case where $\Omega$ is an one point set, a random dynamical system is reduced to a deterministic one. 
\begin{example}
An auto-regressive (AR) model ``$x_{t+1}=Ax_t + v_t$'' is given as a special case of the random dynamical system as follows. Let $\T := \mathbb{Z}$, $\X:=\mathbb{R}^d$. Let $\Omega_0$ be a probability space and $v:\Omega_0\rightarrow \mathcal{X}$ be the Gaussian noise $N(0,\Sigma)$ for some $\Sigma>0$. Let $\Omega:=\Omega_0^\mathbb{Z}$.  We define $\theta_m:\Omega\rightarrow\Omega; (x_n)_n\mapsto(x_{n+m})_n$.  Put $v_m:\Omega\rightarrow\mathbb{R}^d; (x_n)\rightarrow v_m$.  Then, the function 
\begin{align*}
&\Phi(n, \omega, x) \\
&:= A(A(\cdots(Ax+v_0(\omega))+v_1(\omega))+\cdots)+v_{n-1}(\omega))
\end{align*}
is a random dynamical system with respect to $\{\theta_n\}_{n\in\T}$. Therefore, the $t$-th sample $x_t$ determined by the AR model ``$x_{t+1} = Ax_t + v_t$'' is given by $\Phi(t, \cdot, x_0)$.
\end{example}

(reason for putting the following definition)
We remark that random dynamical systems induce dynamical systems on random variables as in the following definition, which plays a main role to define Perron-Frobenius operators for RDS.
\begin{definition}
\label{DS w.r.t. RDS}
Let $\Phi$ be a random dynamical system in $\mathcal{M}$ with respect to $\Theta = \{\theta_t\}_{t\in \T}$. We define $\bm{\varphi}: \T\times M(\Omega,\mathcal{M})\rightarrow M(\Omega,\mathcal{M})$ by $\bm{\varphi}(t,X)(\omega):= \Phi\left(t,\theta_{-t}(\omega),X(\theta_{-t}(\omega))\right)$.
\end{definition}

\subsection{Comparison of Two Random Processes}
\label{ssec:comp_rp}
Here we describe a common method to define a metric for comparing two random processes \cite{Bill99}. A random dynamical system $\Phi$ in $\mathcal{M}$ gives a random processes $\{\Phi(t,\cdot,x)\}_{t\in\T}$ for each $x\in\mathcal{M}$, thus, in particular, this method provides us with a metric for random dynamical systems. 
We give its generalization in Theorem \ref{relation to KMMD} in terms of the operator theoretic method.

Let ${\bf X}:\Omega\rightarrow C^0(\T, \mathcal{X})$ be a stochastic process with continuous paths (for simplicity, we only consider continuous path in the case of $\T = \mathbb{R}$ or $\mathbb{R}_{>0}$). 
Let $\mathcal{L}({\bf X})$ be the law of ${\bf X}$, namely, the push-forward measure ${\bf X}_*P$ on $C^0(\T, \mathcal{X})$. The common strategy to define a metric between stochastic processes is to define the metric between the laws of the stochastic processes by the metric of a probability measure, for example, kernel maximal mean discrepancy (KMMD), Wasserstein distance, and Kullback–Leibler (KL) divergence (cf. \cite{GS02}).

Let $\kappa$ be a positive definite kernel on $C^0(\T, \mathcal{X})$. For a probability measure $\nu$, we denote by $\mu_\nu\in\hilb_\kappa$ the kernel mean embedding of $\nu$, which is given by $\int_{\Omega}\kappa_x\,d\nu(x)\in \mathcal{H}_{\kappa}$.  Then for two stochastic processes ${\bf X}$ and ${\bf Y}$, we can define the metric
\begin{equation}
\left\langle \mu_{\mathcal{L}({\bf X})}, \mu_{\mathcal{L}({\bf Y})}\right\rangle_{\mathcal{H}_{\kappa}}=\int_\Omega\kappa({\bf X}(\cdot, \omega), {\bf Y}(\cdot, \eta)) dP(\omega)dP(\eta). \label{metric for stoch process}
\end{equation}
If a positive definite kernel $k$ on $\mathcal{X}$ is given, we naturally define a positive definite kernel for $g,h\in C^0(\T, \mathcal{X})$ by $\kappa_{k,\mu}(g,h):=\int_{\mathbb{T}} k(f(t), g(t)) d\mu(t)$. 
We give  a generalization of (\ref{metric for stoch process}) in Theorem \ref{relation to KMMD} in the case where the kernel $\kappa=\kappa_{k,\mu}$ for some $k$ and $\mu$. 
%


\section{Perron-Frobenius operators for random dynamical systems}
\label{sec:r2d}

In this section, we define a Perron-Frobenius operator for random dynamical systems, 
 which is a natural generalization of the operator defined in
 \cite{IFI+18} in Subsection~\ref{ssec: PF op on vvRKHS}. vvRKHSs are employed to introduce operator-valued positive definite kernels,  which enables us to incorporate the effects of random variables.  
 These notions 
are generalizations of the corresponding deterministic case (namely, the case where $\Omega$ is an one point set).  
For simplicity, in this section, we assume $\mathbb{T}=\mathbb{R}$ or $\mathbb{Z}$.  

\subsection{A brief review of vector-valued RKHSs}
\label{ssec: vvRKHS}
Let $\bm{\mathcal{X}}$ be a set and $V$ be a Hilbert space. We denote by $\mathcal{B}(V)$ the space of bounded linear operators in $V$. We define a $\mathcal{B}(V)$-valued positive definite kernel ${\bf k}$ as a map ${\bf k}:\mathcal{X}\times\mathcal{X}\rightarrow \mathcal{B}(V)$ satisfying the following two conditions: (1) for any $x,y\in \mathcal{X}$, ${\bf k}(x,y)={\bf k}(y,x)^*$, and (2) for any $r\in \mathbb{N}$, $v_1,\dots,v_r\in V$, and $x_1,\dots,x_r\in \mathcal{X}$, $\sum_{i,j=1}^r\left\langle {\bf k}(x_i,x_j)v_i, v_j\right\rangle_V\ge0$. We note that ``$\mathcal{B}(V)$-valued positive definite kernel'' is equivalent to ``$V$-valued kernel of positive type'' in \cite{CVT06}. We define a linear map ${\bf k}_y: V\rightarrow V^{\mathcal{X}}$ by $({\bf k}_yv)(x)={\bf k}(y, x)v$ for $v\in V$. We note that $\mathbb{C}$-valued positive definite kernel is the equivalent notion to the positive definite kernel in Section \ref{ssec:pf}.

For any $\mathcal{B}(V)$-valued positive definite kernel $k$, it is well known that there uniquely exists a Hilbert space $\hilb_{{\bf k}}$ in $V^{\mathcal{X}}$ such that for any $x\in \mathcal{X}$ and $v\in V$, ${\bf k}_xv\in \hilb_{{\bf k}}$, and for any $h\in\hilb_{{\bf k}}$, $\langle h, {\bf k}_ xv\rangle_{\hilb_{{\bf k}}}=\langle h(x),v\rangle_V$ (see Proposition 2.3 in \cite{CVT06}). We call $\hilb_{{\bf k}}$ the vector-valued reproducing kernel Hilbert space associated with ${\bf k}$ or the $V$-valued reproducing kernel Hilbert space associated with ${\bf k}$. 

We note that we can canonically construct RKHSs from the vvRKHS as in the following proposition:
\begin{proposition}
\label{k from vvRKHS}
For any $v\in V$, the kernel $k_v(x,y):=\langle v, {\bf k}(x,y)v\rangle_V$ is a ($\mathbb{C}$-valued) positive definite kernel. Moreover, the RKHS $\mathcal{H}_{k_v}$ associated with $k_v$ is isomorphic to the closed subspace of $\mathcal{H}_{{\bf k}}$ defined by the closure of the ${\rm span}\{{\bf k}_xv : x\in\mathcal{X}\}$. 
\end{proposition}


\subsection{vvRKHS on random variables}
\label{ssec: vvRKHS on random variables}
Here, we gives vvRKHS for random variables by specifyng vvRKHS described above, 
which plays a central role in the following parts of this paper.

Let $k$ be a bounded and continuous ($\mathbb{C}$-valued) positive definite kernel on a topological space $\mathcal{X}$, and let $(\Omega, P)$ be a probability space with the probability measure $P$.  We denote by $M(\Omega, \mathcal{X})$ the space of $\mathcal{X}$-valued random variables. Now, we define a vvRKHS on $M(\Omega, \mathcal{X})$ in terms of the given kernel $k$.

\begin{definition} 
We define the $\mathcal{B}\left(L^2(\Omega)\right)$-valued positive definite kernel $\mathbf{k}$ on $M(\Omega, \mathcal{X})$ by
\[\left(\mathbf{k}(X, Y)f\right)(\omega) := k(X(\omega),Y(\omega))f(\omega),\]
where 
$X, Y\in M(\Omega, \mathcal{X})$.
\end{definition}
We note that $\mathbf{k}(X,Y)$  induces a bounded linear operator such that $||\mathbf{k}(X,Y)||:=\sup_{\omega\in\Omega}|k(X(\omega),Y(\omega))|$. We may define a natural positive definite kernel for random variables as the expectation of $k(X,Y)$, but, we can mathematically deduce this kernel in our framework: Let $\mathbf{1}\in L^2(\Omega)$ be a constant function valued 1.  We see that we can construct the positive definite kernel $k_{\mathbf{1}}(X,Y):=\mathbb{E}_P[k(X, Y)]$ by Proposition \ref{k from vvRKHS}.


\subsection{Perron-Frobenius Operators for RDS}
\label{ssec: PF op on vvRKHS}
In this subsection, we construct Perron-Frobenius operators for random dynamical systems.  Random dynamical systems include random effects in addition to the structure of dynamical systems.  To this end, we utilize the vvRKHS in Section \ref{ssec: vvRKHS on random variables}, and construct linear operators on the vvRKHS by means of the dynamical systems defined in Definition \ref{DS w.r.t. RDS}. 




First, let $\mathcal{X}$ be a topological space and let ${\bf k}$ be the $\mathcal{B}(L^2(\Omega))$-valued positive definite kernel on $\mathcal{X}$ as in Proposition \ref{ssec: vvRKHS on random variables}.   For any subspace $\mathcal{M}\subset\mathcal{X}$, and random dynamical system $\Phi$ in $\mathcal{M}$, we define
a closed subspace $\hilb_{{\bf k}, \Phi}\subset\hilb_{{\bf k}}$ as the closure of ${\rm span}\left\{{\bf k}_Xh~\big|~X=\Phi(t,\cdot,x), x\in\mathcal{M}, t\in\mathbb{T}, h\in L^2(\Omega)\right\}$.  
\begin{definition}[Perron-Frobenius operators for RDS]
For $t\in\T$, we define the $t$-th {\em Perron-Frobenius operator} for RDS $\Phi$ by the linear operator $K_{\Phi}^t:\hilb_{{\bf k}, \Phi}\rightarrow\hilb_{{\bf k},\Phi}$ with the domain ${\rm span}\left\{{\bf k}_Xh~\big|~X=\Phi(t,\cdot,x), x\in\mathcal{M}, t\in\mathbb{T}, h\in L^2(\Omega)\right\}$ such that $K_{\Phi}^t{\bf k}_Xh={\bf k}_{\bm{\varphi}(t,X)}h$, where $\bm{\varphi}$ is the dynamical system defined in Definition \ref{DS w.r.t. RDS}.
\end{definition}

We give some connections between our operators and other existing operators:

\paragraph{Connection to Perron-Frobenius operators for random processes}
We show that our operator is a generalization of the Koopman operator defined in \cite{CZMM19}, namely, the Perron-Frobenius operator $K_{\Phi}^t$ completely recovers the classical transfer operator in Proposition \ref{random Koopman} below.
Let 
\begin{align}
\iota:\hilb_k\rightarrow\hilb_{\mathbf{k}} \label{def of iota}
\end{align}
be 
$\iota(h)(X):=(h\circ X)$,
and regard $\hilb_k$ as a closed subspace of $\hilb_{\mathbf{k}}$. 
Accurately, at first, we define $\iota(k_x):=\mathbf{k}_x\mathbf{1}$, and then show it induces the well-defined linear map from $\mathcal{H}_k$ to $\mathcal{H}_{\mathbf{k}}$.  
Note that for $g\in \hilb_{\mathbf{k}}$, the adjoint of $\iota$ is defined to be $(\iota^*g)(x)=\mathbb{E}_P[g(x)]$. Here we regard $x\in\mathcal{M}$ as a constant function on $\Omega$, and $g(x)\in L^2(\Omega)$. 
\begin{proposition}\label{random Koopman}
For any $x\in\mathcal{X}$, we have
\[\iota^*K_{\Phi}^{t}\iota(k_x)=\int_\Omega k_{\Phi(t,\omega,x)}dP(\omega).\]
\end{proposition}

\paragraph{Connection to Koopman operators on vvRKHS}
Here, we show the relation between existing operators in vvRKHS and our operator.  In the case where a dynamical system is deterministic ($\Omega$ is an one-point set), discrete-time ($\T=\mathbb{Z}$ or $\mathbb{Z}_{>0}$). Then $K_{\Phi}^1$ is the same one defined in \cite{IFI+18}.  
Let $\mathcal{M}=\mathcal{X}$. In \cite{FK19}, they define the Koopman operator for a discrete-time dynamical system by the linear operator $g\mapsto g\circ f$ for $g\in \hilb_{\mathbf{k}}$, where we put $f:=\varphi(1,\cdot)$. As stated in the following proposition, their operator is given as the adjoint of the Perron-Frobenius operator:
\begin{proposition}\label{adjoint of Koopman}
We have $(K_{\varphi}^1)^*=\mathcal{K}_{{\bf k}}$, where $\mathcal{K}_{{\bf k}}$ is the Koopman operator defined in \cite{FK19} (we give a rigorous definition of the Koopman operator in the proof of this proposition).
\end{proposition}
\section{Metrics on Random Dynamical Systems}
\label{sec:metric}

As in the previous section, we fix a bounded positive definite kernel $k$, and suppose $\mathbb{T}$ is $\mathbb{R}$ or $\mathbb{Z}$.  In this section, we construct a metric to compare two random dynamical systems.  At first, we specify the rigorous definition of the domain where our metric is defined, which we call {\em triples of random dynamical systems} with respect to $k$.  
Then we define the metric on the triples of the random dynamical systems. 
Our metric is given as a positive definite $V$-kernel for some Hilbert space (we specify $V$ later), namely it is a linear operator on $V$. 
When we evaluate this metric with a linear functional, for example the trace, it becomes a usual positive definite kernel. 
Then we see our metric gives a generalization of KMMD for random processes introduced in Section \ref{ssec:comp_rp}, and define two metrics which we use in empirical computation in Section \ref{sec:experiment}.

\subsection{Definition of our metrics}
Let $\nu$ be a Borel measure on $\T$, and let $\hilb_{\rm in}$ and $\hilb_{\rm ob}$ be Hilbert spaces. 
We define triples of a random dynamical system  with respect to $k$ and $\nu$ by
$D=(L, \Phi, \I),$
where $\Phi$ is a random dynamical system on $\mathcal{M}\subset\X$, and $\I:\hilb_{\rm in}\rightarrow\hilb_{\mathbf{k},\Phi}$ and $L:\hilb_{\mathbf{k},\Phi}\rightarrow \hilb_{\rm ob}$  such that $LK_{\Phi}^t\I$ is a Hilbert-Schmidt operator for $t\in\mathbb{T}$, and for any $v\in\hilb_{\rm in}$, the function of $t$ $||LK_{\Phi}^t\I v||_{\hilb_{\rm ob}}\in L^2(\T, \nu)$. 
We call $\I$ and $L$ an {\em initial value operator} and an {\em observable operator}, respectively.
Intuitively, the operator $L$ corresponds to an observable that gives an output at $\hilb_{\rm ob}$, and $\I$ describes an initial condition for data. We denote by
\[\mathscr{T}_k(\hilb_{\rm in}, \hilb_{\rm ob};\nu)\]
the set of the triples of random dynamical systems. 

Now, we give the definition of our metric on random dynamical systems as follows:
\begin{definition}[Metrics for RDS's]
For $i=1,2$, we fix  a triple $D_i:=(L_i, \Phi_i, \I_i)\in\mathscr{T}_k(\hilb_{\rm in}, \hilb_{\rm ob};\nu)$.
For $m\in \mathbb{N}$, $T\in \T$, we define a Hilbert-Schmidt operator by
\begin{align}
\mathscr{K}_k^{(m)}(D_1, D_2):=\bigwedge^m\int_\T\left(L_2K_{\Phi_2}^t\I_2\right)^*L_1K_{\Phi_1}^t\I_1\,d\nu(t),
\end{align}
\vspace*{-5mm}\\
and we define
\begin{align}
\mathfrak{K}_m^T(D_1, D_2):={\rm tr}\left(\mathscr{K}_k^{(m)}(D_1,D_2)\right).
\end{align}
Here, $\wedge^m$ is the $m$-th exterior product (cf.\@ Appendix~\ref{exterior product}).  
\end{definition}
Then we have the following theorem:
\begin{theorem}
\label{main thm}
The kernel $\mathscr{K}_k^{(m)}$ is a $\mathcal{B}\left(\bigwedge^m\hilb_{\rm in}\right)$-valued positive definite kernel on $\mathscr{T}_k(\hilb_{\rm in}, \hilb_{\rm ob}; \nu)$ for each $\nu$.
\end{theorem}
We note that $\mathfrak{K}_m^T$ becomes the positive definite kernel introduced in \cite{IFI+18} in the special case where random dynamical systems are not random but deterministic.

\subsection{Estimation of the metric}
Here we give estimaton formula (\ref{eq:estimation}) and (\ref{eq:estimation2}) below for our metric.  The proof of these formula is given in Appendix \ref{proof of estimation formula}. We use these formula to carry out the numerical experiment in Section \ref{sec:experiment}.
For $i=1,2$, let $\Phi_i$ be a random dynamical system in $\mathcal{M}_i$ and let $x_i^1,\dots,x_i^m\in\mathcal{M}_i$.  In the case of $\hilb_{\rm in}=\mathbb{C}^m$, $\hilb_{\rm ob}=\hilb_{\mathbf{k}}$, $\mathcal{I}_i((a_p)_{p=1}^m):=\sum_{p=1}^m a_p\iota(\phi(x_i^p))$ (here, $\iota$ is defined in (\ref{def of iota})),
we define 
\[\mathfrak{l}_m^T\left((\Phi_1, \bm{x}_1), (\Phi_2, \bm{x}_2)\right)
:=\mathfrak{K}_m^T((\iota^*, \Phi_1, \I_1), (\iota^*, \Phi_2, \I_2)),\]
where $\bm{x}_i:=(x_i^1,\dots,x_i^m)\in\mathcal{M}_i^m$.  Then,  we have the computation formula as follows: 
\begin{equation}
\begin{split}
&\mathfrak{l}_m^T\left((\Phi_1, \bm{x}_1), (\Phi_2, \bm{x}_2)\right)\\
&
=\int\label{eq:estimation}
\det\left(
k\left(\Phi_1(t_p, \omega_p, x_1^p),\,\Phi_2(t_q, \eta_q, x_2^q)\right)\right)_{p,q=1,\dots,m}
\end{split}
\end{equation}
where the variable $(t_1,\dots,t_m, \omega_1,\dots,\omega_m, \eta_1,\dots,\eta_j)$ in the integral runs over $[0,T]^m\times\Omega^m\times\Omega^m$.
In particular, our metric is a generalization of KMMD for an integral type kernel, which is given through the following theorem: 
\begin{theorem}\label{relation to KMMD}
Let 
\[\kappa_T(g,h):=\int_0^Tk(g(t), h(t))d\nu(t)\]
be an integral type kernel on $C^0(\T, \mathcal{X})$, then we have
\begin{align*}
&\mathfrak{l}_1^T\left((\Phi_1, \bm{x}_1), (\Phi_2, \bm{x}_2)\right)= 
\left\langle \mu_{\mathcal{L}\left(\Phi_1(\cdot,\cdot,x_1^1)\right)},\,  \mu_{\mathcal{L}\left(\Phi_2(\cdot,\cdot,x_2^1)\right)}\right\rangle_{\hilb_{\kappa_T}}.
\end{align*}
\end{theorem}
This theorem implies for general $m$, the metric $\mathfrak{l}_m^T$ is a reasonable generalization for two random dynamical systems in the context of KMMD. However,  for higher $m$, 
the formula (\ref{eq:estimation}) needs a heavy calculation. To improve the drawback of (\ref{eq:estimation}), we construct another metric as follows: Let $\hilb_{\rm ob}:=\hilb_{\mathbf{k}}$, and define 
\begin{equation*}
\widetilde{\mathfrak{l}}_m^T((\Phi_1, \bm{x}_1), (\Phi_2, \bm{x}_2))\\
:=\mathfrak{K}_m^T\left(({\rm id}_{\hilb_{\mathbf{k}}}, \Phi_1, \I_1), ({\rm id}_{\hilb_{\mathbf{k}}}, \Phi_2, \I_2)\right).
\end{equation*}
Then we have the following formula:
\begin{align}
&\widetilde{\mathfrak{l}}_m^T\left((\Phi_1, \bm{x}_1), (\Phi_2, \bm{x}_2)\right)\nonumber\\
&=\int
\det\left(
k\left(\Phi_1(t_p, \omega_p, x_1^p),\,\Phi_2(t_q, \omega_q, x_2^q)\right)\right)_{p,q=1,\dots,m}
.\label{eq:estimation2}
\end{align}
where the variable $(t_1,\dots,t_m, \omega_1,\dots,\omega_m)$ in the integral runs over $[0,T]^m\times\Omega^m$.
\section{Connection to Hilbert-Schmidt Independence Criteria}
\label{sec:connection}
In this section, we argue a relation between our operator-valued positive definite kernel and Hilbert-Schmidt independence criteria (HSIC). We first define a independence criterion for random dynamical systems based on the above contexts, and then give its estimator. Our independence criterion measures a pairwise independence of random dynamical systems. Although our metric is constructed in the context of the dynamical system, it can also be used to extract the information about the independence of two systems. This is a main reason to introduce the vvRKHS above.

\subsection{A brief review of Hilbert-Schmidts independence criterion}

We first briefly review HSIC \cite{GBSS05}.  Let $X, Y:\Omega\rightarrow\mathcal{X}$ be two random variables. And, let $k$ be a positive definite kernel on $\mathcal{X}$. Here, we assume that $k$ is characteristic, namely, the kernel mean embedding is injective \cite{SFL10}. Also, we define the {\em cross covariance operator} $C_k(X,Y):\hilb_k\rightarrow\hilb_k$ by
\[C_k(X,Y):=\int_{\Omega}\left(\rule{0pt}{9pt}k_{X(\omega)}-\mu_X\right)\otimes\left(\rule{0pt}{9pt}k_{Y(\omega)}-\mu_Y\right)\,dP(\omega),\]
where $\mu_X, \mu_Y\in\hilb_k$ are the kernel mean embeddings of the lows $X_*P$ of $X$ and $Y_*P$ of $Y$, respectively. We regard any element of $\hilb_k\otimes\hilb_k$ as a bounded linear operator on $\hilb_k$, namely, for any $v\otimes w\in\hilb_k\otimes\hilb_k$, we define $v\otimes w\in\B(\hilb_k)$ by $(v\otimes w)(x):= \langle w, x\rangle v$. 
We note that via the identification, $\hilb_k\otimes\hilb_k$ is equal to the space of Hilbert-Schmidt operators, in particular, the cross-covariance operator $C_k(X,Y)$ is also a Hilbert-Schmidt operator.  
Straightforward computations show that 
\[C_k(X,Y)=\mu_{[XY]}-\mu_X\otimes\mu_Y,\]
where $[XY]:\Omega\rightarrow\mathcal{X}\times\mathcal{X}; \omega\rightarrow (X(\omega), Y(\omega))$. 
The characteristic property of $k$ shows that $C_k(X,Y)=0$ if and only if $X$ and $Y$ are independent. The HSIC is defined to be 
${\rm HSIC}_k(X, Y):=||C_k(X,Y)||_{\rm HS}^{2}:={\rm tr}(C_k(X,Y)^*C_k(X,Y)),$ which is the Hilbert-Schmidt norm of $C_k(X,Y)$. The value ${\rm HSIC}_k(X, Y)$ can be estimated via the evaluation of kernel functions over samples (see \cite[Lemma 1]{GBSS05}).

\subsection{HSIC for random dynamical systems}
Let us consider the independence of random dynamical systems in our context.  For $i=1,2$, let $\Phi_i$ be a random dynamical system on $\mathcal{X}$. 
We fix $x_1,x_2\in\mathcal{X}$, and let $\mathbf{X}:=\{X_t\}_{t\in\T}=\{\Phi_1(t,\cdot, x_1)\}_{t\in\T}$, $\mathbf{Y}:=\{Y_t\}_{t\in\T}=\{\Phi_2(t,\cdot, x_2)\}_{t\in\T}$.  We set $\hilb_{\rm in}:= L^2(\Omega)$ and $\hilb_{\rm ob}:=\hilb_k$.
We define $\I_{x_i}(h):=\mathbf{k}_{x_i}\left(h-\int_\Omega h\right)$. 
Set $D_i:=(\iota_k^*, \Phi_i, \I_{x
_i})\in\mathscr{T}_k(\hilb_{\rm in},\hilb_{\rm ob};\nu_i)$. We define the independent criteria for random dynamical systems by
\begin{align}
&\mathbf{C}_k((\Phi_1,x_1,\nu_1), (\Phi_2, x_2, \nu_2))\nonumber\\
&:={\rm tr}\left(\mathscr{K}^{(1)}_k(D_1,D_1)\mathscr{K}^{(1)}_k(D_2,D_2)\right).
\end{align}

Then, we have the following relation:
\begin{theorem}
\label{relation to HSIC}
We have
\begin{align*}
    &\mathbf{C}_k((\Phi_1,x_1,\nu_1), (\Phi_2, x_2, \nu_2))\\
    &=\int_\T\int_\T C_k(X_s, Y_t)d\nu_1(s)d\nu_2(t).
\end{align*}
\end{theorem}

\begin{figure}[t]
\begin{minipage}{0.49\hsize}
        \centering
\includegraphics[width=1\textwidth]{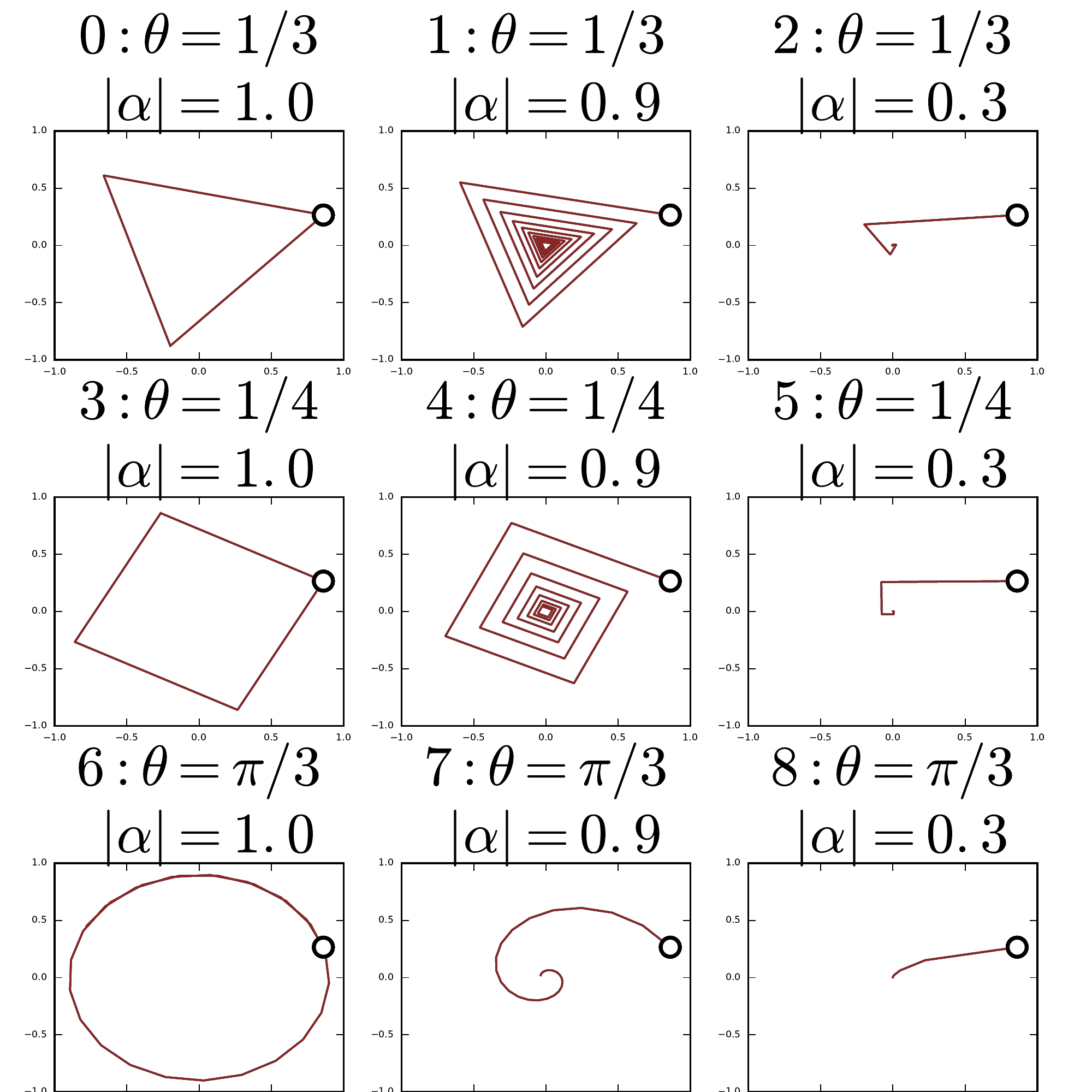}
\subcaption{deterministic}
    \label{fig:data_sigma0}
 \end{minipage}
\begin{minipage}{0.49\hsize}
        \centering
\includegraphics[width=1\textwidth]{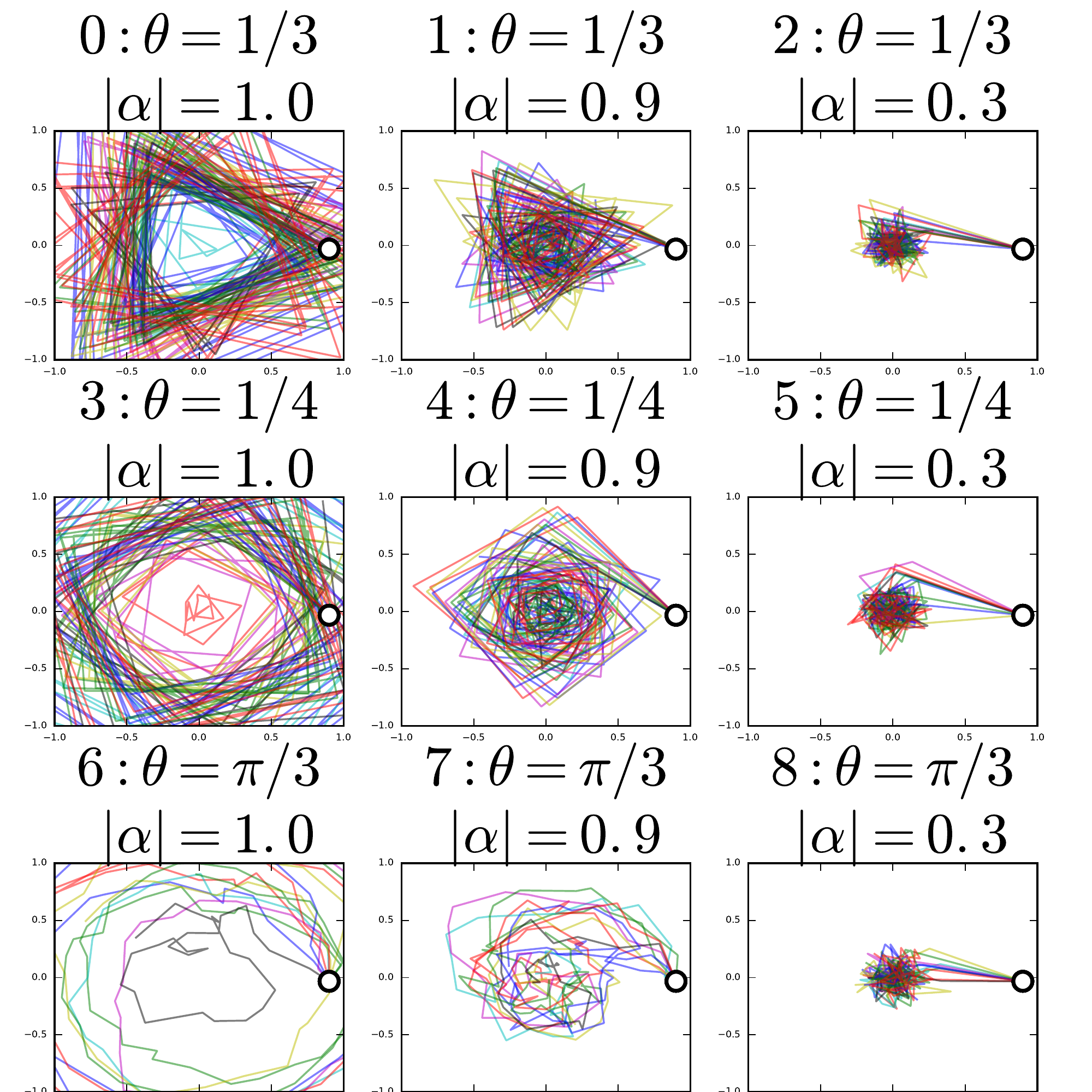}
\subcaption{$\sigma=0.1, N_s=10$}
    \label{fig:data_sigma02}
 \end{minipage}
 \vspace*{-2mm}
 \caption{Orbits of rotation dynamics by (a):just multiplying $\alpha= |\alpha|e^{2\pi i \theta}$ with same initial values, (b): multiplying $\alpha$ and perturb it by Gaussian noise with variance $\sigma^2$.
 }
 \label{fig:data}
\end{figure}

Next we consider the estimation of $\mathbf{C}_k$.
Let $\{X^{(1)}_t\}_{t\in\T}$, $\dots$, $\{X^{(n)}_t\}_{t\in\T}$ and $\{Y^{(1)}_t\}_{t\in\T}$,$\dots$,$\{Y^{(n)}_t\}_{t\in\T}$ be independent $n$ sample passes for $\mathbf{X}$ and $\mathbf{Y}$, respectively. 
Set $\mathbf{Z}:=\mathbf{X}$ or $\mathbf{Y}$, $Z_t^{(p)}:=X_t^{(p)}$ or $ Y_t^{(p)}$, $D:=D_i$, and $\nu:=\nu_i$. We define 
\begin{align}
\widehat{k}_{D}^{(n)}(t;x,y)\nonumber&:=k(x,y)-\frac{1}{n}\sum_{p=1}^nk(x,Z_{t}^{(p)})\nonumber\\
&\hspace{-30pt}- \frac{1}{n}\sum_{i=p}^nk(Z_{t}^{(p)}, y) + \frac{1}{n^2}\sum_{p,q=1}^nk(Z_{t}^{(p)},Z_{t}^{(q)}).
\end{align}
Let $\widehat{G}_{k,D}^{(n)}(t) := \left(\widehat{k}_{D}^{(n)}(Z_t^{(p)}, Z_t^{(q)})\right)_{p,q=1,\dots,n}$ be a matrix of size $n$. By Theorem \ref{relation to HSIC}, and Theorem 1 in \cite{GBSS05}, we have an estimator of $\mathbf{C}_k((\Phi_1, x_1, \nu_1), (\Phi_2, x_2, \nu_2))$  as 
\begin{align}
\label{estimator}
&\widehat{\mathbf{C}}_k((\Phi_1, x_1, \nu_1), (\Phi_2, x_2, \nu_2))\nonumber\\
&:=\frac{1}{(n-1)^2\#S_1\#S_2}\sum_{(s,t)\in S_1\times S_2}{\rm tr}\left(G_{k,D_1}^{(n)}(s)G_{k,D_2}^{(n)}(t)\right),
\end{align}
where $S_1$ and $S_2$ are finite samples according to $\nu_1$ and $\nu_2$, respectively.

\section{Empirical Evaluations}
\label{sec:experiment}
We first empirically illustrate how our metric behaves by using synthetic data from  rotation dynamics with noise on the unit disk in the complex plane in Subsection~\ref{subsec: artificial}, and then evaluate it in the context of the independence test for random processes in Subsection~\ref{ssec:test_eval}.  In the end, we carry out clustering task with real time series data in Subsection \ref{clustering task}.
The codes for generating the results are included in the supplementary.

\begin{figure}[t]
\begin{minipage}{0.49\hsize}
\includegraphics[width=1\textwidth]{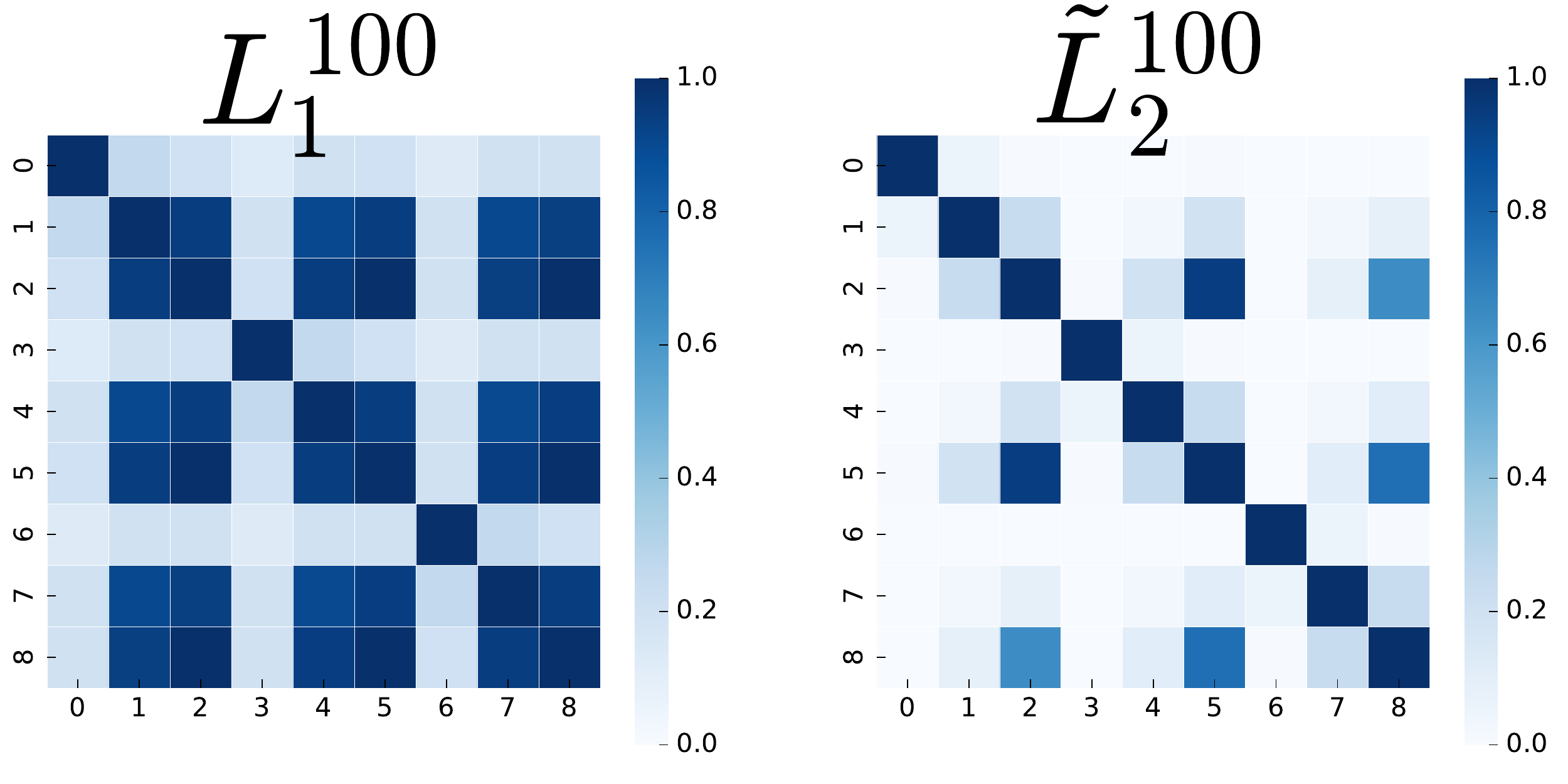}
\subcaption{deterministic}
    \label{fig:A_sigma0}
 \end{minipage}
\begin{minipage}{0.49\hsize}
        \centering
\includegraphics[width=1\textwidth]{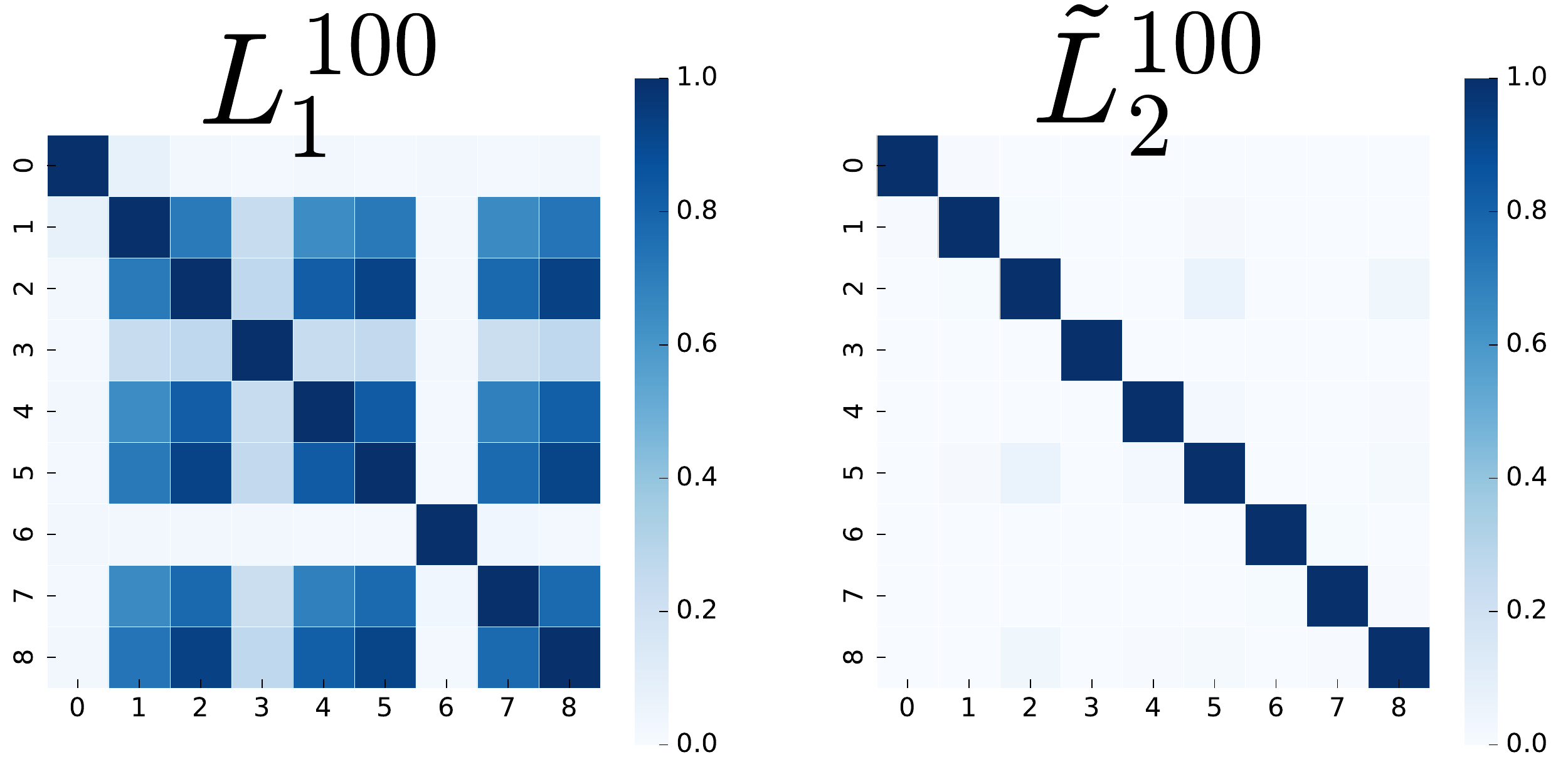}
\subcaption{$\sigma=0.1, N_s=1$}
    \label{fig:A_sigma01_Ns1}
 \end{minipage}
\begin{minipage}{0.49\hsize}
        \centering
\includegraphics[width=1\textwidth]{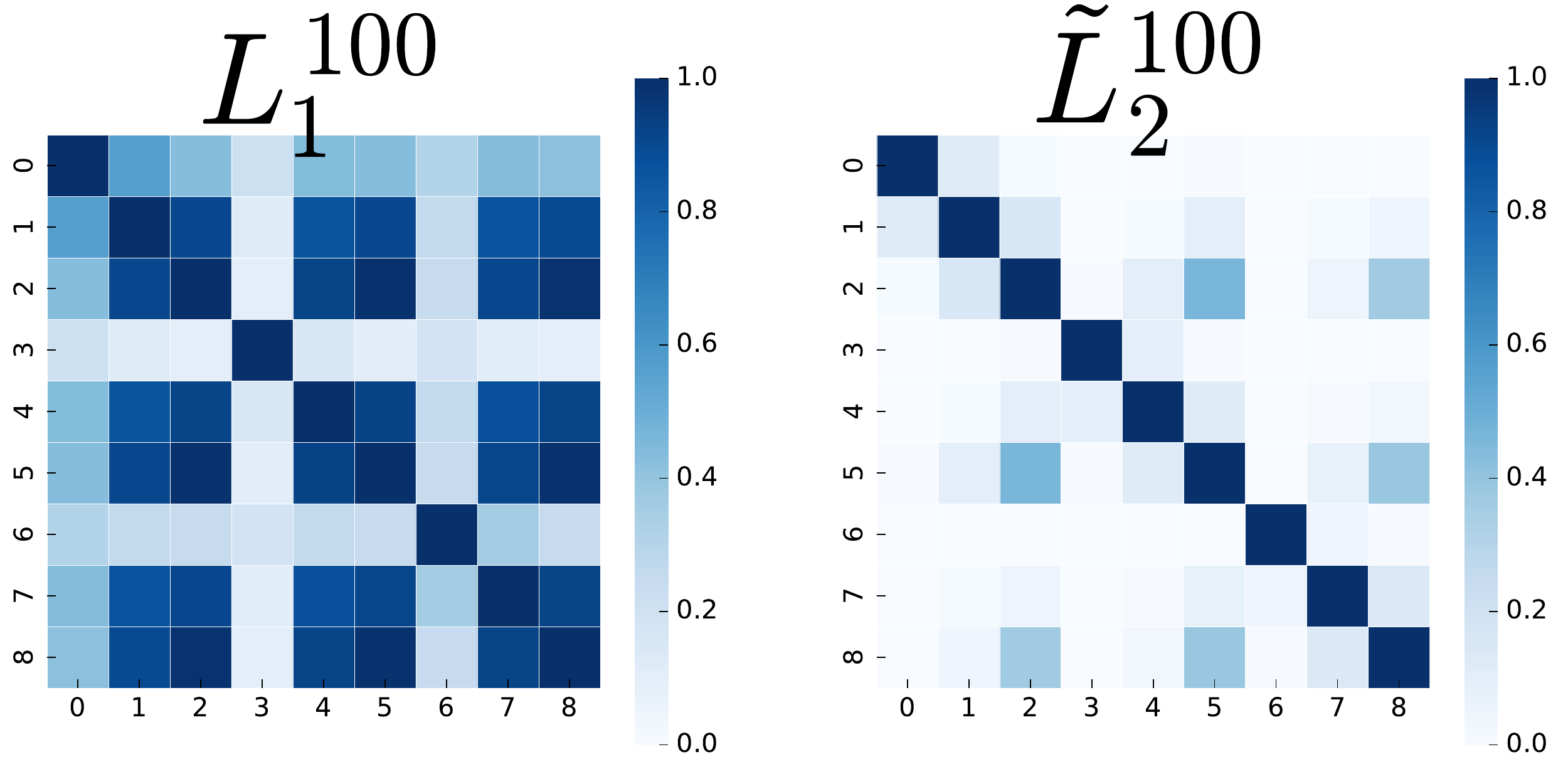}
\subcaption{$\sigma=0.1, N_s=10$}
    \label{fig:A_sigma01_Ns10}
 \end{minipage}
 \begin{minipage}{0.49\hsize}
        \centering
\includegraphics[width=1\textwidth]{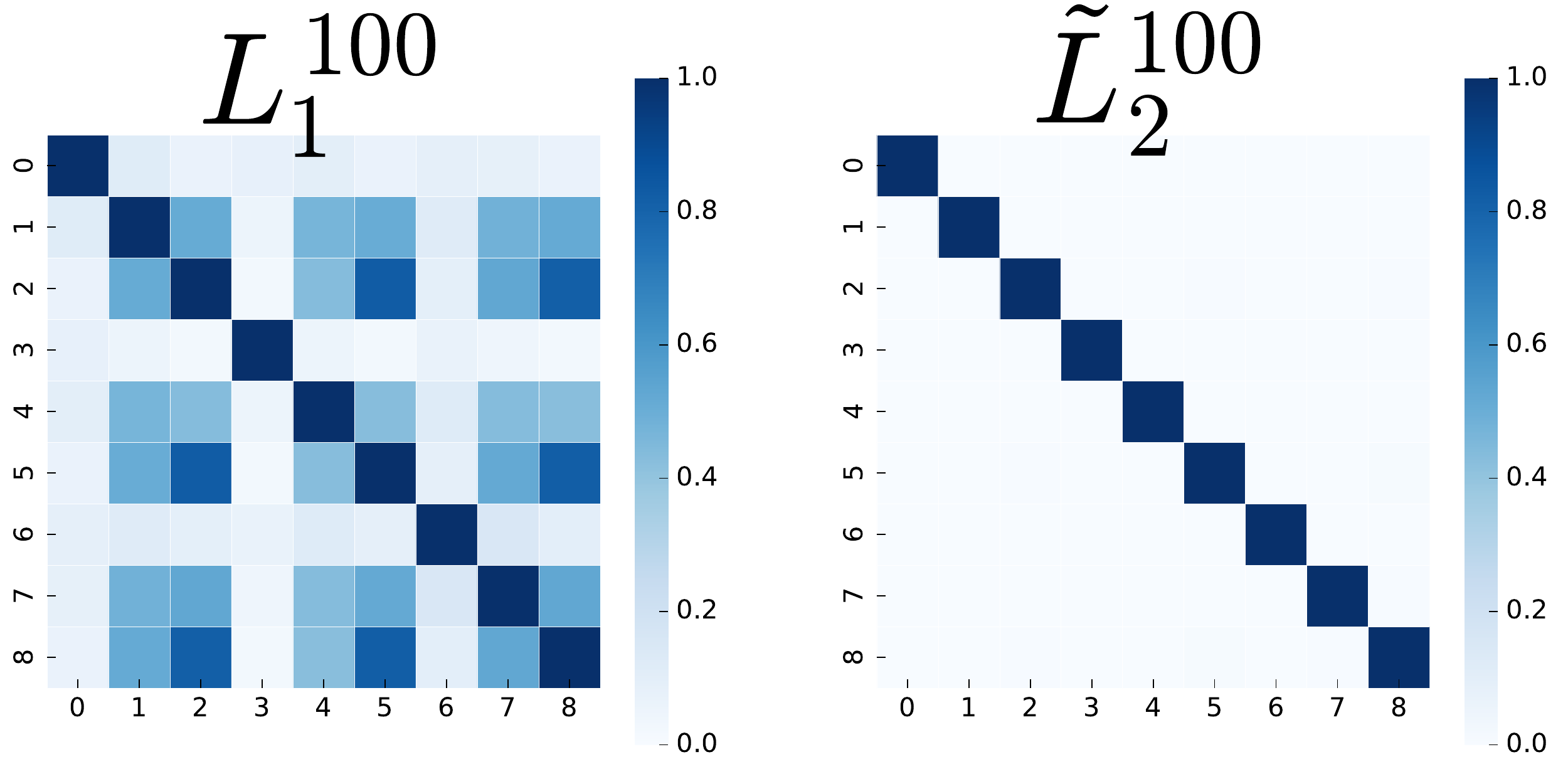}
\subcaption{$\sigma=0.5, N_s=10$}
    \label{fig:A_sigma05_Ns10}
 \end{minipage}
 \vspace*{-2mm}
 \caption{Discrimination results of various $(\sigma, N_s)$ pairs.
 Vertical and horizontal axes correspond to the dynamics in Figure \ref{fig:data}.
 In this experiment, the Gaussian kernel $k(z, w) := \exp{(-|z-w|^2/2)}$ for $z,w\in \mathbb{C}$ is used.
 }
 \label{fig:As}
\end{figure}

\subsection{Illustrative Example of Rotation Dynamics with Noise}
\label{subsec: artificial}
We used synthetic data from the rotation  dynamics $R_\alpha:z\mapsto\alpha z$ for $\alpha\in\mathbb{C}$ with $|\alpha|\leq 1$ with a noise $\epsilon_t$ on the unit disk $\{z\in \mathbb{C};|z|\le 1\}$ in the complex plane, where the variance of the noise is  $\sigma^2$, i.e. $\epsilon_t \sim \mathcal{N}(0, \sigma^2)$ (i.i.d.).
We prepared $3 \times 3$ combination of parameters $\alpha = |\alpha|e^{2 \pi i \theta}$ with $|\alpha| \in \{1, 0.9, 0.3 \}$ and $\theta \in \{ 1/3, 1/4, \pi/3 \}$.
The graphs in Figure \ref{fig:data} show the deterministic case and 10 independent paths for the case with $\sigma=0$ from the identical initial condition $z_0$ with $|z_0|=0.9$, where the lines of different colors show different sample paths. 
Then, we calculated the normalized variant of our metric defined by 
\begin{align}
&L_m^T ((\Phi_1, \bm{x}_1), (\Phi_2, \bm{x}_2))
\nonumber\\
&=
\lim_{\epsilon \to +0}
\frac{|\mathfrak{l}_m^T\left((\Phi_1, \bm{x}_1), (\Phi_2, \bm{x}_2)\right) + \epsilon |^2}
{\displaystyle\prod_{i=1}^2|\mathfrak{l}_m^T\left((\Phi_i, \bm{x}_i), (\Phi_i, \bm{x}_i)\right) + \epsilon |}
\label{eq:L}
\end{align}
with an empirical approximation of $\mathfrak{l}_m^T$ defined in \eqref{eq:estimation}. We also define $\widetilde{L}_m^T$ by replacing $\mathfrak{l}_m^T$ with $\widetilde{l}_m^T$, whose empirical estimation is given by \eqref{eq:estimation2}. 
We computed $L_1^T$ and $\widetilde{L}_2^T$ here.
The graphs in Figure \ref{fig:As} show numerical results for some cases.
As can be seen in (b), if the number of samples $N_s$ 
is rather small compared with the strength of the noise $\epsilon_t$, we observe that $L_1^T$ only captures rough similarity and $\widetilde{L}_2^T$ judges all dynamics are different. However, it tend to improved in (c) as the number of samples $N_s$ get larger. Also, $\widetilde{L}_2^T$ gives similar results with the deterministic case.
As for (d) where the noise level is stronger, $\widetilde{L}_2^T$ again seems to judge all dynamics are different.

\subsection{Independence between Two Time-series Data}
\label{ssec:test_eval}
\begin{figure}[t]
\centering
\includegraphics[width=.48\textwidth]{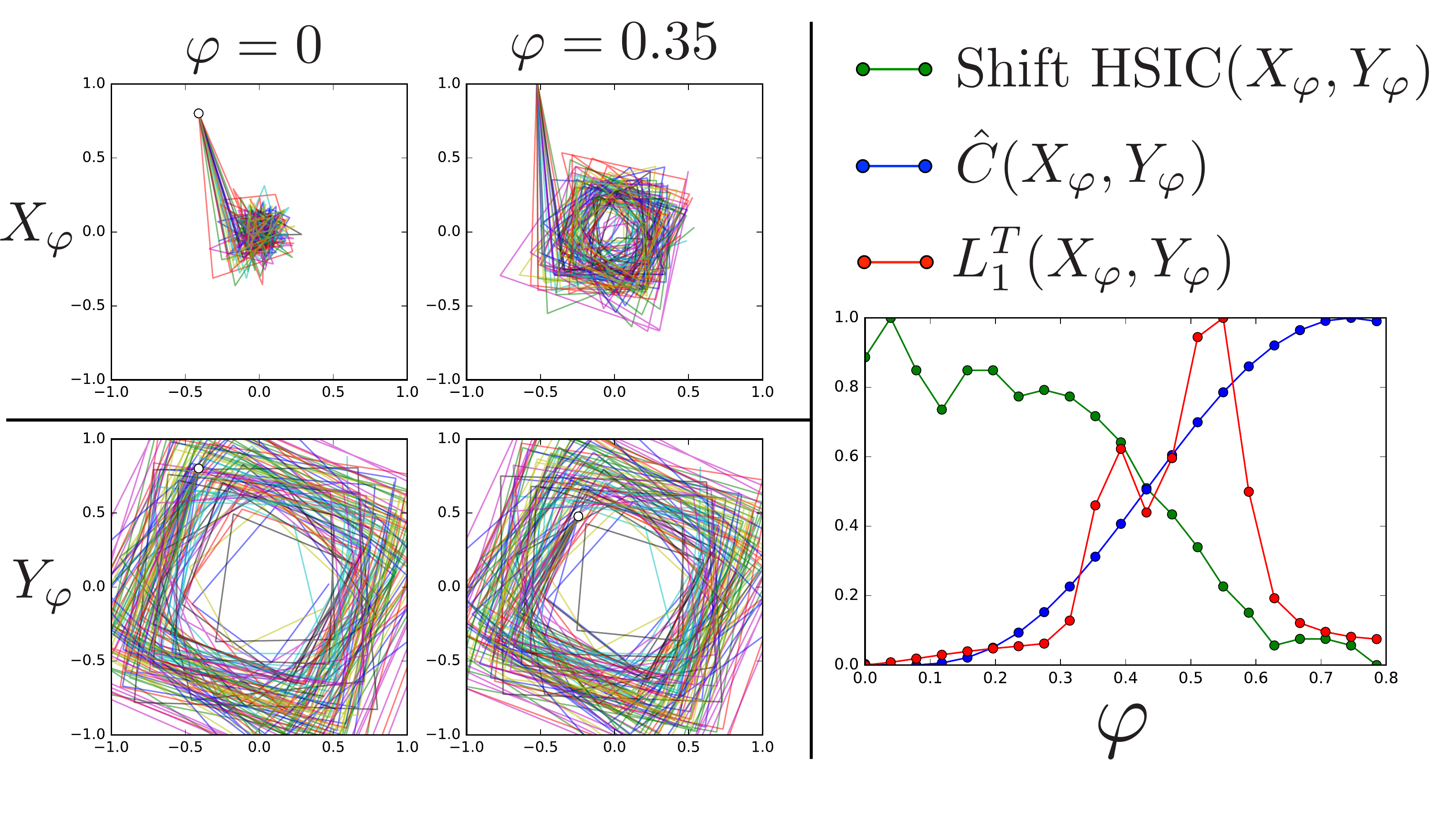}
 \vspace*{-9mm}
 \caption{
 Left: Examples of two random series $X$ and $Y$ independently generated by \eqref{Xseries} and \eqref{Yseries}, and those linear combinations $X_\varphi$ and $Y_\varphi$ with $\varphi=0.35$.
 Right: plots for shift HSIC, $\hat{\bm C}_k(X_\varphi, Y_\varphi)$ and $L_m^T$ with $m=1$ and $T=20$ for $\varphi \in [0, \pi/4]$.
Each plot is normalized as max=1 min=0.
 }
 \label{fig:XYdata}
\vspace*{-2mm}
\end{figure}

We empirically evaluated the effectiveness of our metric as an independence criterion. For this purpose, we first generated $N_s=20$ pairs of complex valued time-series data with total time $T=20$ by
\begin{align}
&
x_{t+1} = 0.3 e^{\frac{2 \pi i}{3}} x_t (1- 0.01 x_t)
+ \epsilon_t^X,
\label{Xseries}
\\
&
y_{t+1} = 0.9 e^{\frac{2 \pi i}{4}} y_t (1- 0.01 y_t)
+ \epsilon_t^Y,
\label{Yseries}
\end{align}
where noises are sampled from complex normal distribution: $\epsilon_t^X, \epsilon_t^Y \sim \mathcal{CN}(0, 0.1^2)$.
We denote by $X = [x_1, x_2, \dots, x_{T=20}]$ and $Y = [y_1, y_2, \dots, y_{T=20}]$ the generated sequences. Then, we created different data pairs as
\begin{align}
&X_\varphi := X \cos \varphi + Y \sin \varphi,
\\
&Y_\varphi := - X\sin \varphi + Y\cos \varphi
.
\end{align}
By the definitions, $X_\varphi$ and $Y_\varphi$ are independent for $\varphi =0$, and correlated for $\varphi \neq 0$.
The two subplots in $\varphi=0$ column in the graphs of the left-hand side in Figure~\ref{fig:XYdata} show 10 independently generated samples for $X$ and $Y$, where the lines of different colors show different sample paths. 
On the other hand, samples of $X_{\varphi=0.35}$ and $Y_{\varphi=0.35}$ share the square shapes and seem to be correlated.

And, the graph of the right-hand side in Figure~\ref{fig:XYdata} shows the calculated $\hat{\bm C}_k(X_\varphi, Y_\varphi)$ (blue) and $L_m^T(X_\varphi, Y_\varphi)$ (red) with these sample paths. 
As a baseline, we also calculated the mean value of shift HSICs  among $20 \times 20$ paths (green) with head=10 and tail=15 \cite{CG14}. 
All values are normalized so that the maximum value is 1.0 and the minimum value is 0.  

As one can see, the both graphs for Shift HSIC and $\hat{\bm C}_k(X_\varphi, Y_\varphi)$ are monotone along the independence of the random processes. It is consistent to their ability to measure the independence of the systems.  In contrast,  the metric $1-L_1^T$ has a peak between 0.3 and 0.5.  It reflects that $L_m^T$ detects not the independence of the random behavior but the difference of the dynamics, which can be seen in Section \ref{subsec: artificial} as well.


\subsection{Clustering real data}
\label{clustering task}
\begin{figure}[t]
\centering
 \includegraphics[width=0.45\textwidth]{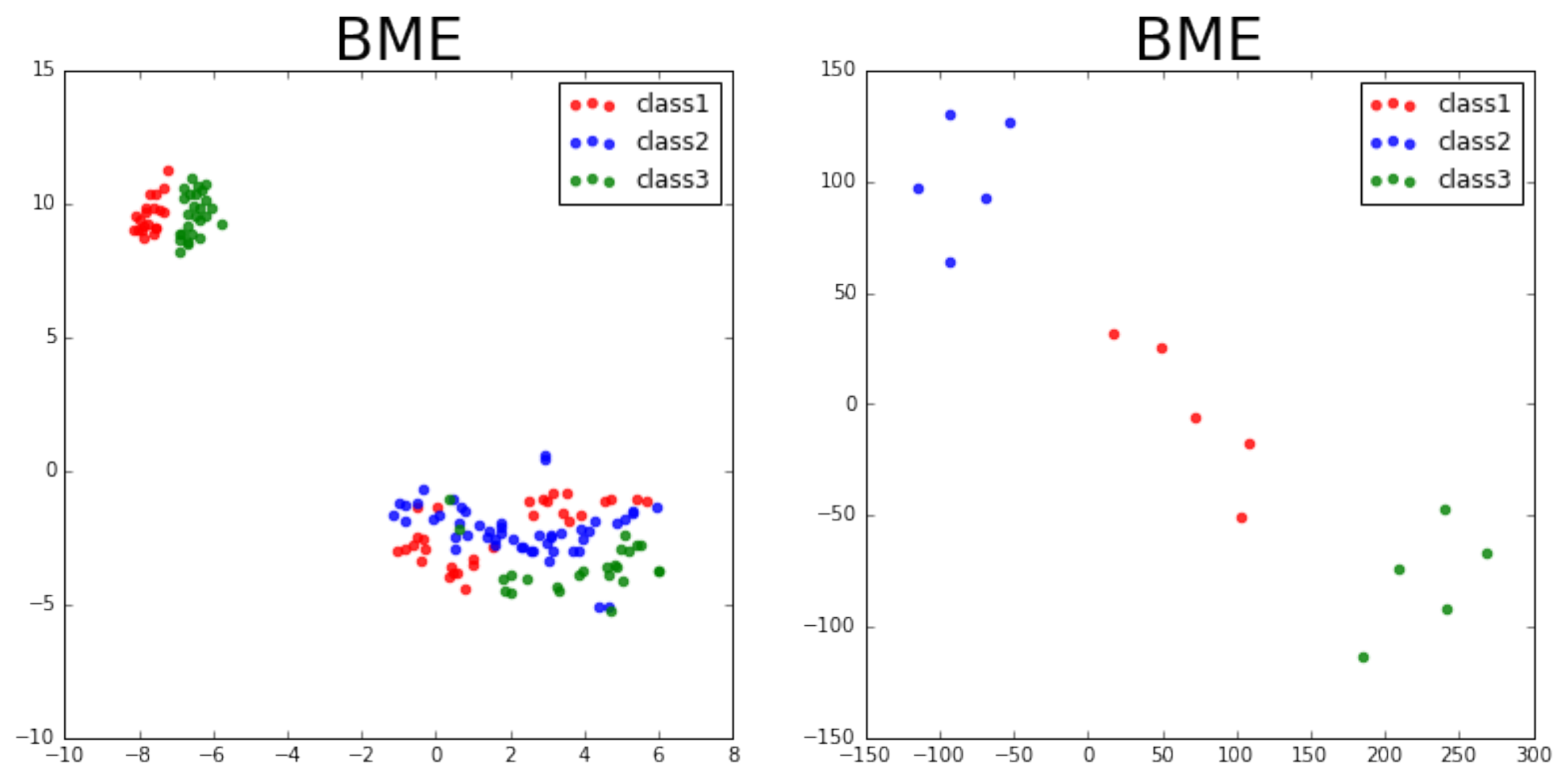}\\
  \includegraphics[width=0.45\textwidth]{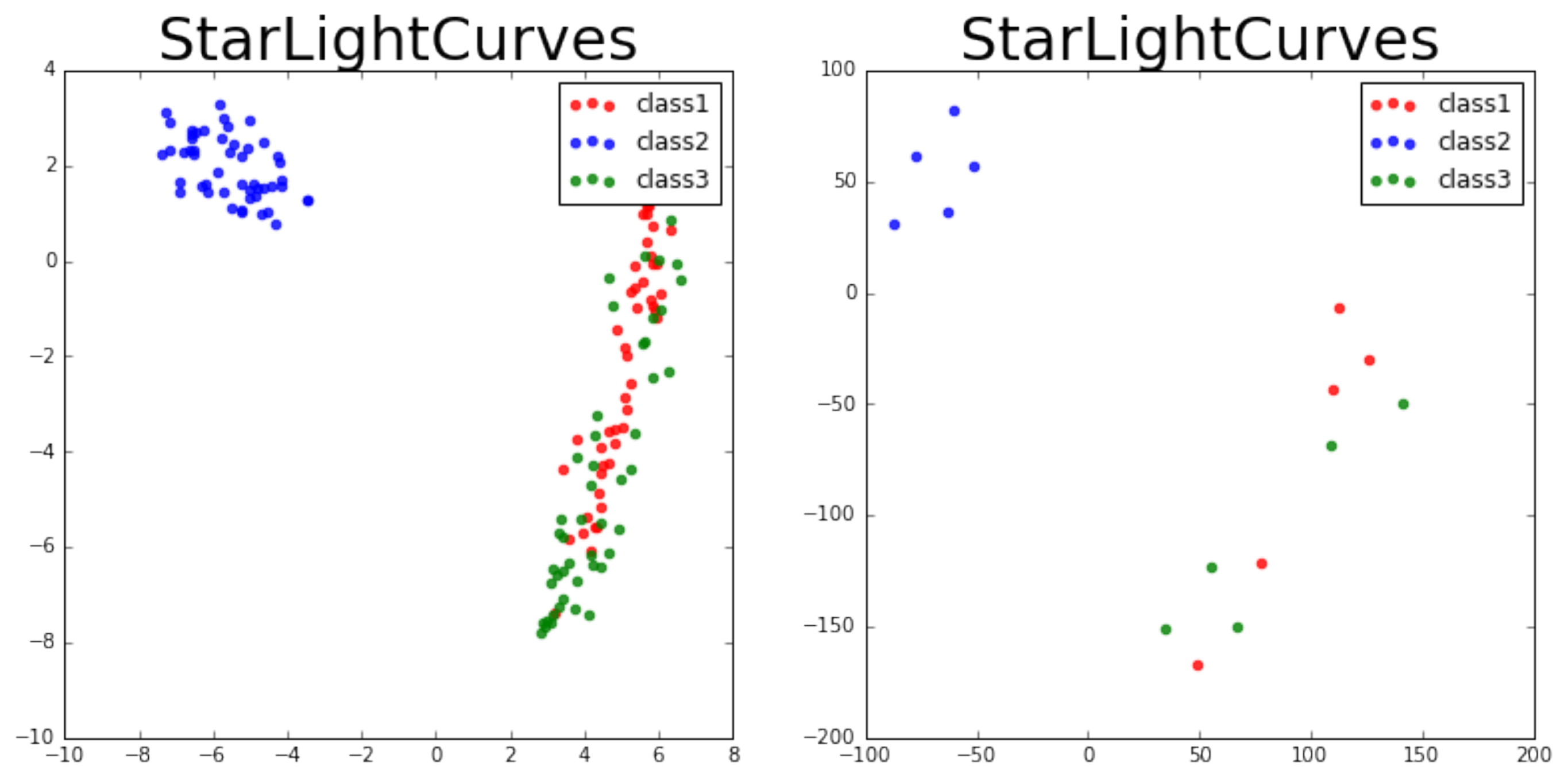}\\
  \ 
 \vspace*{-2mm}
 \caption{
 Left column: tSNE with perplexity=50 by using $1-A_1^T$.
 Right column:
 tSNE with perplexity=5 by using $1-L_1^T$.
 data.
 }
 \label{fig:BME}
\vspace*{-5mm}
\end{figure}
As an experiment using real-world data, we used UCR Archive in 2018 \cite{UCRArchive2018}.
We show results for TEST data in BME, and StarLightCurves
in Figure \ref{fig:BME}.
Each data is composed of $N_c$-class pairs of $N_\text{paths}$ paths with time-length $T$.
To apply our method and compare it to the existing one, we first take $n_\text{paths} < N_\text{paths}$ paths from each class and divide it to $n_\text{pairs}$ pairs of $n_\text{sub-paths}$ paths, and calculate $(N_c\cdot n_\text{pairs}) \times (N_c\cdot n_\text{pairs})$ distance matrix using Eq.~\eqref{eq:L}, i.e. $1-L_1^T$.
We used $n_\text{paths} = 50, n_\text{pairs}=5$, i.e. $n_\text{sub-paths}=10$ in BME and StarLightCurves data.
We also calculated $(N_c\cdot n_\text{paths}) \times (N_c\cdot n_\text{paths})$ distance matrix using $1- A_1^T$ proposed in \cite{IFI+18} and applied tSNE \cite{Maaten08}.
To reduce the computational costs, we took skipped time series $X^\text{skipped} = [x_1, x_{s+1}, x_{2s+1}, \dots]$.
We used $s_\text{BME}=1$, ad  $s_\text{StarLightCurves} = 100$. 

In the left side of the Figure \ref{fig:BME}, we suppose the time series data purely comes from deterministic dynamical systems and do not take random effects into account.  On the other hand, in the right hand side, we suppose it is generated by random systems. Since our metric includes an averaging process, it could cluster BME data perfectly.  However, regarding StarLightCurves data, it seems not to be able to clearly separate two classes (red and green). This would imply the latent dynamics behind the two classes are quite similar.

\section{Conclusions}
\label{sec:concl}

In this paper, we developed a general framework for constructing metrics on random nonlinear dynamical systems with the Perron-Frobenius operators in vvRKHSs. vvRKHSs were employed to design mathematically manageable metrics and also to introduce $\mathcal{B}\left(L^2(\Omega)\right)$-valued kernels, which enables us to handle the randomness in systems. 
Our metric is an extension of the existing metrics for deterministic systems. Also, we generalized the Hilbert-Schmidt independence criteria to time series data. We empirically showed the effectiveness of our metric using examples of rotation dynamics with noise in the unit disk in the complex plane and clustering tasks for real time series data.  We also evaluated our metric in the context of the independence test for random processes. 

\appendix
\newcommand{\kk}{\mathbf{k}}
\section{Proofs}
\subsection{Proposition \ref{random Koopman}}
For any $X\in M(\Omega,\mathcal{X})$ and $v\in L^2(\Omega)$, we claim that 
$\iota^*(\kk_X v)(x)=\int_{\Omega}k_{X(\omega)}v(\omega)dP(\omega)$.
In fact, denote by $\alpha$ the right hand side of this claim, then, by straightforward computations, we have
\begin{align}
\langle k_x, \iota^*(\kk_X v)\rangle_{\hilb_k} 
&=\langle k_x, \alpha\rangle_{\hilb_k}, \label{siki 1}
\end{align}
which proves the claim.
Since $\iota(k_x)=\kk_{x}1$, 
for any $t\in \T$ and $x\in\mathcal{M}$, we have $K_{\Phi}^t\iota(k_x)=\kk_{\bm{\varphi}(t,x)}1$. By combining this with (\ref{siki 1}), we see that $\iota^*K_{\Phi}^t\iota(k_x)=\int_\Omega k_{\Phi(t,\omega,x)}dP(\omega)$.\hfill$\blacksquare$

\subsection{Proposition \ref{adjoint of Koopman}}
Set $f:=\varphi(1,\cdot)$.  The definition of the Koopman operator $\mathcal{K}_{\mathbf{k}}$ for vvRKHS is given as follows:
$\mathcal{K}_{\mathbf{k}}:\hilb_{\mathbf{k}}\rightarrow\hilb_{\mathbf{k}}$ is a linear operator with domain $\mathscr{D}(\mathcal{K}_{\mathbf{k}}):=\left\{h\in\hilb_{\mathbf{k}}~\big|~h\circ f \in \hilb_{\mathbf{k}}\right\}$ such that for any $h \in \mathscr{\mathcal{K}_{\mathbf{k}}}$, 
$\mathcal{K}_{\bm{k}}h = h\circ f.$
Then we see that $(K_{\varphi}^1)^* = \mathcal{K}_\kk$ since  for any $v\in V$, $x\in \mathcal{X}$, and $h\in\mathscr{D}(\mathcal{K}_f)$, 
\begin{align*}
\langle K_{\varphi}^1\bm{k}_xv, h\rangle_{\hilb_{\bm{k}}} &=\langle h(f(x)), v\rangle_{\hilb_{\bm{k}}}=\langle \bm{k}_xv, \mathcal{K}_{\bm{k}} h\rangle_{\hilb_{\bm{k}}}.\\
&&\hfill\blacksquare
\end{align*}

\subsection{Theorem \ref{main thm}}
We denote by $L^2(\T,\nu; V)$ the space of $L^2$-integrable $V$-valued functions with respect to the measure $\nu$, where $V$ is any Hilbert space. Let $Q_i(t):=L_iK_{\bm{\varphi},W_i}\I_i$.
Let $R_{D_i}:\hilb_{\rm in}\rightarrow L^2(\T,\nu; \hilb_{\rm ob}); v\mapsto [t\mapsto Q_i(t)v]$. Then we see that the adjoint operator of $R_{D_i}$ is given by
\[R_{D_i}^*h = \int_{\T}Q_i(t)^*h(t)d\nu(t).\]
Therefore, we see that 
$\mathscr{K}_k^{(1)}(D_1, D_2) = R_{D_2}^*R_{D_1}.$
For general $m$, let $R_{D_i,m}:= \wedge^m R_{D_i}$. Then we see that $\mathscr{K}_k^{(m)}(D_1, D_2)=R_{D_2,m}^*R_{D_1,m}$. Therefore, we have $\mathscr{K}_k^{(m)}(D_1, D_2)=\mathscr{K}_k^{(m)}(D_2, D_1)^*$, and for $D_1,\dots, D_r\in\mathscr{T}_k(\hilb_{\rm in}, \hilb_{\rm ob};\nu)$, and $v_1\dots v_r\in\wedge^m\hilb_{\rm in}$,
\[\sum_{i,j=1}^r\langle \mathscr{K}_k^{(m)}(D_i, D_j)v_i, v_j\rangle_{\wedge^m\hilb_{\rm in}}=\big|\big|\sum_{i=1}^rR_{D_i}v_i\big|\big|_{\wedge^m\hilb_{\rm in}}^2\ge0,\]
namely, $\mathscr{K}_k^{(m)}$ is a $\mathcal{B}(\wedge^m\hilb_{\rm in})$-valued positive definite kernel.\hfill$\blacksquare$

\subsection{Theorem \ref{relation to KMMD}}
This theorem follows from the formula (\ref{eq:estimation}) and the definition of the kernel mean embedding.

\subsection{Theorem \ref{relation to HSIC}}
For $i=1,2$, put $E_i=(\Phi_i, x_i, \nu_i)$.  Set $Q_i(t):=\iota^*K_{\Phi}^t\I_{x_i}$. Then, we see that
\begin{align*}
&\mathbf{C}_k(E_1, E_2)
=\int_{\T}\int_{\T}||Q_2(t)Q_1(s)^*||_{\rm HS}^2d\nu_1(s)d\nu_2(t).
\end{align*}
Thus, it suffices to show that $Q_2(t)Q_1(s)^* = C_k(X_s, Y_t)$. Set $\Phi=\Phi_i$, $x=x_i$, $Q(t)=Q_i(t)$, and let $\tilde{h}:=h-\int_\Omega h(\omega)dP(\omega)$. A straight computation shows that 
\begin{align*}
Q(t)h
&=\int_{\Omega}(k_{Z_t(\omega)}-\mu_{Z_t})h(\omega)dP(\omega),
\end{align*}
where $Z_t=X_t$ or $Y_t$. Thus we see that $Q(t)^*v(\omega)=\langle v, k_{Z_t(\omega)}-\mu_{Z_t}\rangle_{\hilb_k}$. Therefore, we have
\[Q_2(t)Q_1(s)^*v = \int_{\Omega}(k_{Y_t(\omega)}-\mu_{Y_t})\langle v, k_{X_s(\omega)}-\mu_{X_s}\rangle_{\hilb_k}dP(\omega),\]
namely, $Q_2(t)Q_1(s)^* = C_k(X_s, Y_t)$.\hfill $\blacksquare$

\newpage
{\small
\bibliography{metric_rds_nips19}
\bibliographystyle{plain}}
\newpage

\section{Exterior product of Hilbert spaces}
\label{exterior product}
Let $H$ be a Hilbert space with inner product $\langle\cdot,\cdot\rangle$.  Let $H^{\otimes m}$ be a $m$-tensor product as an abstract complex linear space.  Then for $x_1\otimes\dots x_m, y_1\otimes\dots y_m\in H^{\otimes m}$, 
\[\langle x_1\otimes\dots x_m, y_1\otimes\dots y_m\rangle_\otimes:=\prod_{i=1}^m\langle x_i, y_i\rangle\]
induces an inner product on $H^{\otimes m}$.  We denote by $\widehat{\otimes}^mH$ the completion via the norm induced by the inner product $\langle\cdot, \cdot\rangle_\otimes$.  

We define a linear operator $\mathscr{E}: \widehat{\otimes}^mH\rightarrow \widehat{\otimes}^mH$ by
\[\mathscr{E}(x_1\otimes x_m):=\sum_{\sigma\in\mathfrak{S}_m}{\rm sgn}(\sigma)x_{\sigma(1)}\otimes\cdots\otimes x_{\sigma(m)},\]
where $\mathfrak{S}_m$ is the $m$-th symmetric group, and ${\rm sgn}:\mathfrak{S}_m\rightarrow\{\pm 1\}$ is the sign homomorphism.
We define the {\em $m$-th exterior product} of $H$ by
\[\bigwedge^mH:=\mathscr{E}\left(\widehat{\otimes}^mH\right).\]
For $x_1,\dots, x_m\in H$, we also define 
\begin{align*}x_1\wedge\dots x_m:=\mathscr{E}(x_1\otimes\dots \otimes x_m) 
\end{align*}
The inner product on $\bigwedge^mH$ is described as
\begin{align}\langle x_1\wedge\dots \wedge x_m, y_1\wedge\dots \wedge y_m\rangle_{\bigwedge^mH}=\det(\langle x_i, y_j\rangle)_{i,j=1,\dots,m}.\label{inner product for wedge}
\end{align}
We note that there exists an isomorphism
\begin{align*}
&\bigoplus_{r+s=m}\bigwedge^rH\widehat{\otimes}\bigwedge^sH'\\
&\cong\bigwedge^m(H\oplus H'); \sum_{r+s=m}x_r\otimes y_s\mapsto \sum_{r+s=m}x_r\wedge y_s.
\end{align*}

Let $L: H\rightarrow H'$ be a linear operator. Then $L$ induces a linear operator $\widehat{\otimes}^m L: \widehat{\otimes}^mH\rightarrow \widehat{\otimes}^mH'$ defined by $\widehat{\otimes}^mL(x_1\otimes \dots x_m):=Lx_1\otimes\dots\otimes Lx_m$. The operator  $\widehat{\otimes}^m L$ induces an operator on $\bigwedge^m H$, namely,$\widehat{\otimes}^m L\left(\bigwedge^m H\right)\subset\bigwedge^m H'$,   and we define
\[\bigwedge^m L:=\left.\widehat{\otimes}^m L\right|_{\bigwedge^m H}.\]

\section{Proof of (\ref{eq:estimation}) and (\ref{eq:estimation2})}
\label{proof of estimation formula}
The proof of (\ref{eq:estimation}) and (\ref{eq:estimation2}) are done in a similar way, so we only give the proof of (\ref{eq:estimation}) here.  By the proof of Lemma 3.7 in \cite{IFI+18} (see also Appendix E of \cite{IFI+18}), we see the following formula:
\begin{align*}
&\mathfrak{l}^T_m((\Phi_1,x_1),(\Phi_2, x_2))\\
&=\int_{\mathbb{T}^m}
\left\langle \iota^*K_{\Phi_1}^{t_1}\iota\phi(x_1^1)\wedge\dots\wedge \iota^*K_{\Phi_1}^{t_m}\iota\phi(x_1^m),\right.\\
&\hspace{40pt}\,\left.\iota^*K_{\Phi_2}^{t_1}\iota\phi(x_2^1)\wedge\dots\wedge \iota^*K_{\Phi_2}^{t_m}\iota\phi(x_2^m)\right\rangle_{\wedge^m\mathcal{H}_k}.
\end{align*}
Thus, by Proposition \ref{random Koopman} and the definition of the inner product for $\wedge^m\mathcal{H}_k$ (\ref{inner product for wedge}), we have the formula (\ref{eq:estimation}).

\section{Other experiments in Subsection 6.2}
\begin{figure*}[t]
\begin{minipage}{1.\hsize}
\centering
 \includegraphics[width=1.\textwidth]{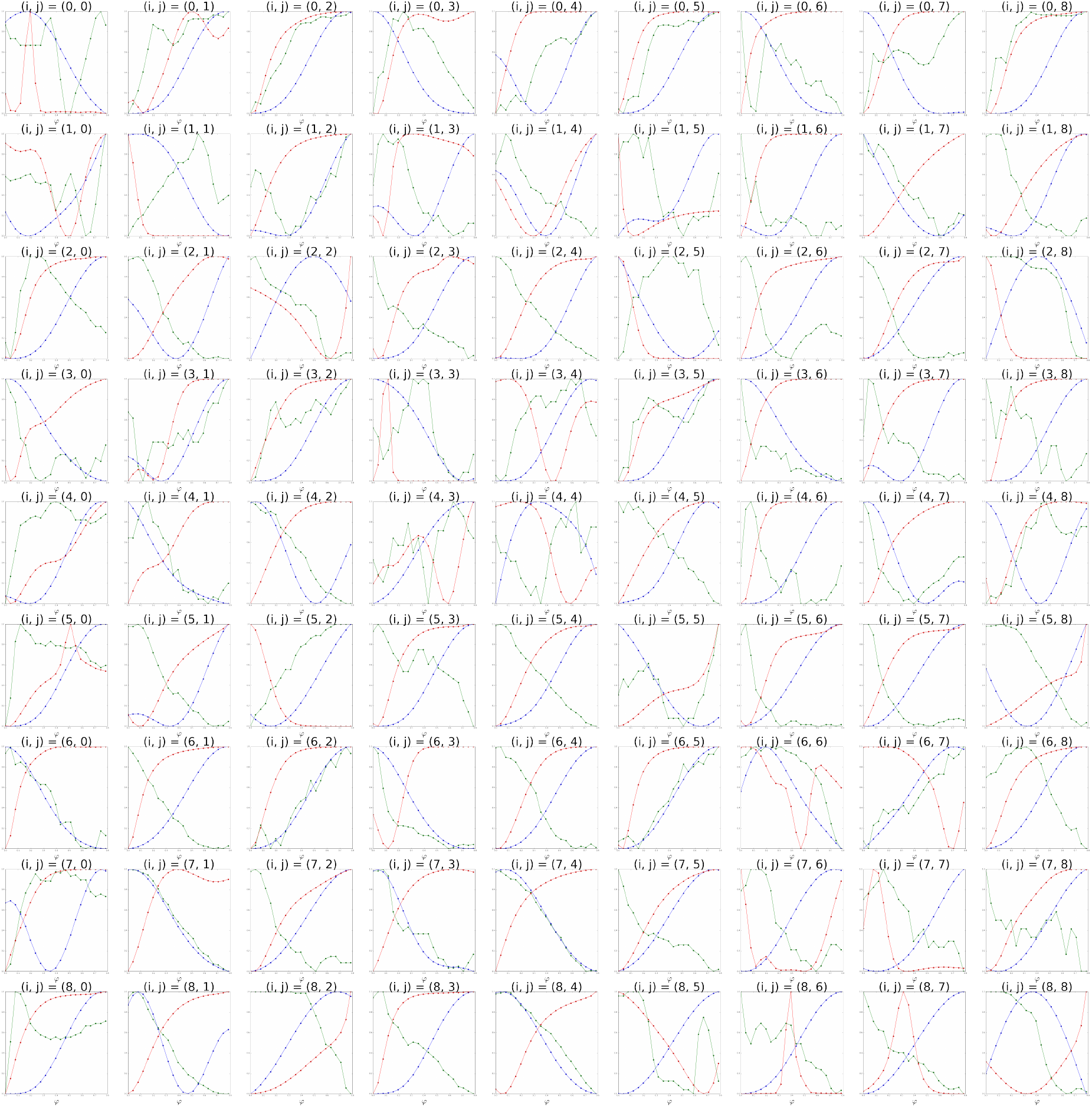}
 \vspace*{-2mm}
 \caption{
 Plots of normalized shiftHSIC$(X^{(i)}_\varphi, X^{(j)}_\varphi)$ (green) $\hat{\bm C}_k(X^{(i)}_\varphi, X^{(j)}_\varphi)$ (blue) and $L_1^T(X^{(i)}_\varphi, X^{(j)}_\varphi)$ (red) for $\varphi \in [0, \pi/4]$.
 }
 \label{fig:allC}
\end{minipage}
\end{figure*}

In Subsection 6.2, we only show an experiment for independence criterion with a particular pair of random dynamical series $X$ and $Y$.
Here, we show results of experiments using $9 \times 9$ combinations of $X^{(i)}$ and $X^{(j)}$ defined by
\begin{align}
x_{t+1}^{(i)}
=
|\alpha^{(i)}|e^{2 \pi i \theta^{(i)}}
x_t^{(i)}
(1- 0.01 x_t^{(i)})
+
\epsilon_t 
,
\notag
\end{align}
where $\epsilon_t \sim \mathcal{CN}(0, 0.1^2)$ and each pair of parameters $(|\alpha^{(i)}|, \varphi^{(i)})$ is defined by the following table.
\begin{align}
\left. \begin{array}{l||l|l|l|l|l|l|l|l|l|}
i & 0 & 1 & 2 & 3 & 4 & 5 & 6 & 7 & 8\\ \hline
|\alpha^{(i)}| & 
1 & 0.9 & 0.3 &
1 & 0.9 & 0.3 &
1 & 0.9 & 0.3 
\\ \hline
\theta(i) & 
\frac{1}{3} & \frac{1}{3}  & \frac{1}{3}  &
\frac{1}{4} & \frac{1}{4} & \frac{1}{4} &
\frac{\pi}{3} & \frac{\pi}{3} & \frac{\pi}{3}
\\ \hline
\end{array} \right.
 \notag
\end{align}
We pick up two parameter indices $i$ and $j$, and sample $N_s = 10$ independent paths with total time $T=10$.
From these sample paths, we define another pairs of paths
\begin{align}
X^{(i)}_\varphi &= 
X^{(i)} \cos \varphi + X^{(j)} \sin \varphi \notag \\
X^{(j)}_\varphi &= 
-X^{(i)} \sin \varphi + X^{(j)} \cos \varphi, \notag
\end{align}
and calculate shiftHSIC$(X^{(i)}_\varphi, X^{(j)}_\varphi)$, $\hat{\bm C}_k (X^{(i)}_\varphi, X^{(j)}_\varphi)$ and $L_1^T(X^{(i)}_\varphi, X^{(j)}_\varphi)$ with $k$ as the Gaussian kernel.
We plot these values in Figure \ref{fig:allC} for $\varphi \in [0, \pi/4]$ with all combinations of $i, j$.
\end{document}